\documentclass[runningheads]{llncs}
\usepackage[misc]{ifsym}

\usepackage{graphicx}
\usepackage[utf8]{inputenc} 
\usepackage[T1]{fontenc}    
\usepackage{hyperref}       
\usepackage{url}            
\usepackage{booktabs}       
\usepackage{amsfonts}       
\usepackage{nicefrac}       
\usepackage{microtype}      

\usepackage{amsmath}
\usepackage{tabularx}
\usepackage{graphicx}
\usepackage{color}
\usepackage{algorithm}
\usepackage[noend]{algpseudocode}
\usepackage{multicol}
\usepackage{multirow}
\usepackage{upgreek}
\usepackage{diagbox}
\usepackage{array}

\graphicspath{{images/}}
 
\newcolumntype{Y}{>{\centering\arraybackslash}X}

\newlength{\mylength}
\makeatletter
\newcommand{\mycfs}[1]{%
  \normalsize
  \@defaultunits\mylength=#1pt\relax\@nnil
  \edef\@tempa{{\strip@pt\mylength}}%
  \ifx\protect\@typeset@protect
     \edef\@currsize{\noexpand\mycfs\@tempa}
  \fi
  \mylength=1.2\mylength
  \edef\@tempa{\@tempa{\strip@pt\mylength}}%
  \expandafter\fontsize\@tempa
  \selectfont
}
\makeatother

\begin{document}
\title{Fast and Accurate Importance Weighting for Correcting Sample Bias}
%
%
\author{Antoine de Mathelin\inst{1, 2}{\Letter} \and
Francois Deheeger\inst{1} \and
Mathilde Mougeot\inst{2} \and
Nicolas Vayatis\inst{2}
}

\authorrunning{A. de Mathelin et al.}

\institute{Manufacture Française des Pneumatiques Michelin, Clermont-Ferrand, France \\ \email{\{antoine.de-mathelin-de-papigny, francois.deheeger\}@michelin.com} \and 
Université Paris-Saclay, CNRS, ENS Paris-Saclay, Centre Borelli, Gif-sur-Yvette, France \email{\{mathilde.mougeot, nicolas.vayatis\}@ens-paris-saclay.fr}}

\toctitle{Fast and Accurate Importance Weighting for Correcting Sample Bias}
\tocauthor{Antoine~de~Mathelin}

\maketitle              

\begin{abstract}
    Bias in datasets can be very detrimental for appropriate statistical estimation. In response to this problem, importance weighting methods have been developed to match any biased distribution to its corresponding target unbiased distribution. The seminal Kernel Mean Matching (KMM) method is, nowadays, still considered as state of the art in this research field. However, one of the main drawbacks of this method is the computational burden for large datasets. Building on previous works by Huang et al. (2007) and de Mathelin et al. (2021), we derive a novel importance weighting algorithm which scales to large datasets by using a neural network to predict the instance weights. We show, on multiple public datasets, under various sample biases, that our proposed approach drastically reduces the computational time on large dataset while maintaining similar sample bias correction performance compared to other importance weighting methods. The proposed approach appears to be the only one able to give relevant reweighting in a reasonable time for large dataset with up to two million data.
\end{abstract}

\section{Introduction}
\label{intro}

The most common assumption in a traditional learning scenario is that training data are independently and identically distributed (iid) and drawn from the same distribution as the target data. However, in real cases, the training dataset often appears to be biased with respect to the target dataset. This happens in particular in medical applications, when, for example, the age distribution of the patients does not match the distribution of the overall population. In product design, predictive models of product performances may be biased by the large amount of data corresponding to outdated products. For both previous cases, it often happens that the learner has access to the unbiased distribution, either because it is known from an external source (the age distribution in the whole population is known) or because he has access to an unbiased dataset (a sample of data of the recent products).

In this paper we assume that the learner owns a sample drawn from a source biased distribution $p_s(x, y)$ as well as a sample coming from the target marginal distribution $p_t(x)$ such that $p_s(x) \neq p_t(x)$. Our goal is to estimate $p_t(y|x)$ or $p_t(y)$ where $y$ is the variable of interest (e.g. patient survival expectancy for a clinical model or product performances for product design). Given the bias on the marginals, the estimation of $p_t(y|x)$ on the target domain will be biased as well.

To correct this type of sample bias, importance weighting methods can be used. These methods seek to reweight the source data to debias the marginals by looking for the weights corresponding to $w(x) = p_t(x) / p_s(x)$. A successful non-parametric method in this field is the Kernel Mean matching (KMM) method \cite{Huang2007KMM} which reweights the sources in order to minimize the MMD distance between the reweighted sources and the targets \cite{Gretton2012MMD}. Although KMM is one of the first non-parametric method developed to handle sample bias, it is still used nowadays in modern sample bias correction methods for deep learning \cite{Fang2020DeepImportanceWeighting} or for deriving two-sample hypothesis testing under sample bias \cite{bellot2021kernelTwoSampleTest}. KMM solves a quadratic problem for the minimization of the MMD with as many parameters as the number of source instances. Thus, when the number of source data is large, one faces a computational burden because of the large kernel matrix to compute. Some methods have proposed to reduce the problem in batch and to perform a KMM on each of them \cite{chandra2016BatchKMM}, \cite{Miao2015EnsembleKMM}. This lightens the memory, but the computational time remains important as the number of KMM sub-problems to compute increases with the number of data. Other importance weighting methods reduce the number of parameters to be optimized by linking the weights of the source instances by a parametric function as done for KLIEP \cite{Sugiyama2007KLIEP} and ULSIF \cite{kanamori2009ULSIF}. These two methods propose to write each of the weights as a linear combination of kernels centered on target points. Thus, the number of parameters is fixed (in general by selecting a hundred centers in the target domain), however, the computational cost of the pairwise distance calculations between the centers and all the source data still remains. Moreover, this large matrix of pairwise distances is used in the resolution of the gradient descent algorithm which slows the optimization. A last method, NearestNeighborsWeighting (NNW) consists in computing the weights of the source instances according to their number of target nearest neighbors \cite{loog2012NearestNeighborsWeighting}. This heuristic, not relying on the minimization of a distance between distributions, is quite efficient, and its computation time is generally less than its KMM counterpart. However, the search for the nearest neighbors requires the computation of pairwise distances between source and target data and despite the optimization algorithms of type KDTree \cite{friedman197KDTree} and BallTree \cite{omohundro1989BallTree}, the method encounters computational burden for datasets with many instances and features.

Finally, all these algorithms rely on hyper-parameters to be tuned. The choice of the kernel and its bandwidth for the KLIEP and KMM methods are very important, as well as the number of nearest neighbors to consider for NNW. To choose these parameters, a cross-validation procedure using an unsupervised metric (which does not require the $y$ data on the target domain) is mainly used such as the J-score for KLIEP \cite{Sugiyama2007KLIEP}, the normalized mean squared error (NMSE) between the actual and estimated density ratios for KMM \cite{Miao2013AutoTuningKMM}, an information criterion for ULSIF \cite{kanamori2009ULSIF}, or any divergence metric between distributions such as the linear discrepancy or the domain classifier divergence \cite{BenDavid2006DATheory}, \cite{deMathelin2021HandlingTireDesign}, \cite{Mansour2009DATheory}. This hyper-parameter selection procedure, necessary to use these methods in practice, adds to the computational time.

Considering the drawbacks of the previous mentioned methods, we propose, in this paper, a new importance weighting algorithm that scales to large datasets. Our goal is to obtain the same level of performance than KMM but with less computational time. To do so, we propose to minimize the objective of KMM, i.e. the MMD, by a batch gradient descent to avoid the memory burden of the huge kernel matrix. However, it should be underlined that optimizing the weights of the source instances individually brings no complexity gain since each source weight is only updated in its corresponding batch. Assuming $w(x) = p_t(x) / p_s(x)$ continuous and regular, for two close source points $x_1 \simeq x_2$, the weights will be similar $w(x_1) \simeq w(x_2)$. Consequently, we propose to optimize at each batch the parameters $\theta$ of a parametric and continuous function $W_{\theta}(x)$ in order to minimize the empirical MMD on the batch. Inspired from recent works of weighting adversarial neural network (WANN) \cite{deMathelin2020WANN}, this function $W_{\theta}$ is chosen as a neural network. The advantage of the networks is the fast update of the $\theta$ parameters by backpropagation of the gradient through the layers which is highly parallelizable
\cite{lecun2015deeplearning}. This avoids working with huge matrices of pairwise kernel as done in the KLIEP algorithm. We show on several datasets that this approach allows to obtain importance weighting at least as efficient as KMM in a drastically reduced time. The source code of the experiments is publicly available on GitHub\footnote{\url{https://github.com/antoinedemathelin/Importance-Weighting-Network}}.

\smallskip

\noindent Our contributions can be listed as follows:
\begin{itemize}
    \item We derive a fast and scalable importance weighting algorithm. This is achieved by using batch gradient descent optimizing the MMD and by parameterizing the weights by a neural network.
    \item The developed algorithm optimizes the kernel parameters of the MMD in the gradient-descent optimization and thus avoid a time consuming CV process to select it.
\end{itemize}

\section{Problem Setting and Proposed Approach}
\label{wann}

\subsection{Learning scenario}

Given an input space $\mathcal{X} \in \mathbb{R}^p$ of dimension $p>0$ and an output space $\mathcal{Y} \in \mathbb{R}^q$ with $q > 0$, we consider the sample bias scenario in which the learner has access to a source sample $\mathcal{S} = \{ (x_1, y_1), ..., (x_m, y_m)\} \subset \mathcal{X} \times \mathcal{Y}$ drawn iid from a source distribution $p_s(x, y)$ on $\mathcal{X} \times \mathcal{Y}$ and a target sample $\mathcal{T} = \{ x_1, ..., x_n)\} \subset \mathcal{X}$ drawn iid according to a target distribution $p_t(x)$ on $\mathcal{X}$. We suppose that $p_s(x) \neq p_t(x)$ and that $p_t(x)$ is absolutely continuous with respect to $p_s(x)$. Finally, we make the covariate-shift assumption \cite{Bickel2009CovariateShift} which states that the conditional probabilities of $y|x$ remain unchanged for the two distributions: $p_s(y|x) = p_t(y|x)$.

\subsection{MMD}

Let's consider $\phi_{\sigma}: \mathcal{X} \to \mathcal{F}_{\sigma}$ with $\mathcal{F}_{\sigma}$ the RKHS of Gaussian kernel $k_{\sigma}$ such that $\forall x, x' \in \mathcal{X}, \; k_{\sigma}(x, x') = \langle  \phi_{\sigma}(x) , \phi_{\sigma}(x') \rangle = \exp(- \sigma ||x - x'||^2)$ with $\sigma > 0$. The Maximum Mean Discrepancy (MMD) between the source and target distributions is defined as follows:

\begin{equation}
    \text{MMD}_{\sigma}(p_s(x), p_t(x)) =  \left| \left| \mathop{\mathbb{E}}_{x \sim p_s(x)}[\phi_{\sigma}(x)] - \mathop{\mathbb{E}}_{x \sim p_t(x)}[\phi_{\sigma}(x)] \right| \right|
\end{equation}

The MMD is a distance characterizing how close are the two marginal distributions $p_s(x), p_t(x)$. As we consider a Gaussian kernel, $\text{MMD}_{\sigma}=0$ if and only if $p_s(x) = p_t(x)$ \cite{Gretton2012MMD}.

As our goal is to correct the sample bias between the source and target distributions with importance weighting, we aim at finding the weights $w(x) \in \mathbb{R}_+$ that solve the following optimization problem:

\begin{equation}
\label{optim}
\begin{split}
    &\min_{w: \mathcal{X} \to \mathbb{R}_+} \left| \left| \mathop{\mathbb{E}}_{x \sim p_s(x)}[w(x)\phi_{\sigma}(x)] - \mathop{\mathbb{E}}_{x \sim p_t(x)}[\phi_{\sigma}(x)] \right| \right|^2 \\& \text{subject to} \; w(x)\geq0 \, \forall x \in \mathcal{X} \; \text{and} \mathop{\mathbb{E}}_{x \sim p_s(x)} [w(x)] = 1
\end{split}
\end{equation}

As we consider $p_t(x)$ absolutely continuous with respect to $p_s(x)$, the solution of the optimization problem (\ref{optim}) is the density ratio $w(x) = p_t(x) / p_s(x)$ \cite{Huang2007KMM}.

In practice, we only have access to samples $\{x_1, .., x_m \}$ and $\{x'_1, .., x'_n \}$ respectively drawn according to both distributions $p_s(x)$ and $p_t(x)$, we then consider the empirical formulation of the previous optimization problem (\ref{optim}) which is written:

\begin{equation}
\label{optim_emp}
\begin{split}
    &\min_{w \in \mathbb{R}^m} \frac{1}{m^2} \sum_{i, j}^{m} w_i w_j k_{\sigma}(x_i, x_j) + \frac{1}{n^2} \sum_{i, j}^{n} k_{\sigma}(x'_i, x'_j) - \frac{2}{n m}  \sum_{i}^{m} \sum_j^n w_i k_{\sigma}(x_i, x'_j) \\
    &\text{subject to} \; w_i\geq0 \, \forall i \in [|1, m|] \; \text{and} \; \frac{1}{m} \sum_{i}^{m} w_i = 1
\end{split}
\end{equation}

\subsection{Importance Weighting Network}
\label{iwn}

The optimization problem (\ref{optim_emp}) is a quadratic optimization problem which can be solved by gradient descent. However computing the MMD requires to compute a kernel matrix of size $\mathcal{O}((n+m)^2)$ which can cause memory burden. We propose, in this paper, to compute the MMD on small batches of size $B$. At each batch, we impose the constraints on the weights by taking their absolute values and dividing them by their sum. It has been shown that, although self-normalizing the weights creates a biased estimation of the MMD, the estimator is asymptotically unbiased, with the bias decreasing at a rate of $\mathcal{O}(1/B)$ \cite{Diesendruck2019ImportanceWeightedGenerativeNetwork}, \cite{martino2017SelfNormalizeWeights}.

To obtain a fast update of all weights at each batch, we parameterized the weights through a neural network $W_{\theta}: \mathcal{X} \to \mathbb{R}$ such that $w_i = W_{\theta}(x_i)$ for each $i \in [|1, m|]$. In this way, at each batch, the parameters $\theta$ are updated and then all the parameters $w_i$ are updated with them. Notice that the MMD estimation produced by the batch of size $B$ is approaching the true MMD at a strong rate of $\mathcal{O}(1/\sqrt{B})$ \cite{Gretton2012MMD} which comforts the idea that the update of $\theta$ at each batch will be in favor of finding the optimal weights for all $x_i$.

It should be stressed that the MMD quantity depends on $\sigma$ wich corresponds to the kernel bandwidth. In the seminal paper of KMM \cite{Huang2007KMM} the choice of $\sigma$ is not clearly motivated, but a method proposed by KLIEP \cite{Sugiyama2007KLIEP} consists in choosing between several predefined $\sigma$ and compute the optimal weights for each $\sigma$ value. The value which provides the best matching of the target distribution is finally selected. This type of selection is time consuming and requires fixing a pre-selection of $\sigma$ values.

We propose, instead, to optimize the $\sigma$ parameter at the same time as the weights. Inspired by the works on MMD-GAN \cite{Li2017MMDGAN}, the kernel parameter $\sigma$ is modified at each batch in order to maximize the MMD. The idea behind this choice of implementation is to increase the discriminative power of the MMD and thus reduce the risk of estimating, from finite samples, that the source and target distributions are the same when this is not the case. By maximizing over $\sigma$, we end with an alternate gradient descent-ascent algorithm, where we aim at finding a saddle point. The final optimization formulation can be written as follows:

\begin{equation}
\begin{split}
    \max_{\sigma} \min_{\theta} & \; \frac{\sum_{i, j}^{B} |W_{\theta}(x_i) W_{\theta}(x_j)| k_{\sigma}(x_i, x_j)}{\sum_{i, j}^{B} |W_{\theta}(x_i) W_{\theta}(x_j)|}   \\ & + \frac{1}{B^2} \sum_{i, j}^{B} k_{\sigma}(x'_i, x'_j) \\ & - \frac{2 \sum_{i}^{B} \sum_j^B |W_{\theta}(x_i)| k_{\sigma}(x_i, x'_j)}{B \sum_{i}^{B} |W_{\theta}(x_i)|}  
\end{split}
\end{equation}

We therefore introduce the Importance Weighting Network (IWN) which searches for the saddle points that solve the above optimization problem (cf Algorithm \ref{algo-iwn}).

\begin{algorithm}[t!]
	\caption{\textbf{Importance Weighting Network}}\label{algo-iwn}
	\begin{algorithmic}
	    \State \textbf{Inputs:} Source and target datasets $\mathcal{S}_{\mathcal{X}}, \mathcal{T}$, initial bandwidth $\sigma$, batch size $B$, neural network $W_{\theta}$, learning rate $\nu$
	    \State \textbf{Initialization:} Fit $W_{\theta}$ with loss $\mathcal{L} = \sum_i || W_{\theta}(x_i) -1 ||^2$
	    \While {stopping criterion is not reached}
		\State Take batches $\{x_1, ..., x_B \} \subset \mathcal{S}_{\mathcal{X}}$ and $\{x'_1, ..., x'_B \} \subset \mathcal{T}$
		\State \textbf{Forward propagation}
		\State $w_i \leftarrow |W_{\theta}(x_i)| / \sum_j^B |W_{\theta}(x_j)| \; \forall \, x_i$
		\State $\text{MMD}_{\sigma, \theta} = \sum_{i, j}^{B} w_i w_j k_{\sigma}(x_i, x_j) + \frac{1}{B^2} \sum_{i, j}^{n} k_{\sigma}(x'_i, x'_j) - \frac{2}{B}  \sum_{i}^{B} \sum_j^B w_i k_{\sigma}(x_i, x'_j)$
		\State \textbf{Backward propagation} 
		\State $\theta \leftarrow \theta - \nu \nabla_{\theta} \text{MMD}_{\sigma, \theta}$
		\State $\sigma \leftarrow \sigma + \nu \nabla_{\sigma} \text{MMD}_{\sigma, \theta}$
		\EndWhile
	\end{algorithmic}
\end{algorithm}

\section{Related work}
\label{related-work}

\textbf{Instance-based domain adaptation}. Our work is in line with instance-based unsupervised transfer learning or domain adaptation \cite{Pan2010survey}. Most of the instance-based methods have already been introduced previously as KMM, KLIEP, ULSIF and RULSIF \cite{Huang2007KMM}, \cite{Sugiyama2007KLIEP}, \cite{kanamori2009ULSIF}, \cite{yamada2011RULSIF}. All of these methods aim to compute the source weights which minimize a distance between the input distributions like the MMD or the Kullback-Leibler. Other methods have also proposed to take into account the model used to estimate $y$ using appropriate metrics such as the discrepancy \cite{Cortes2014DAregression}, \cite{Mansour2009DATheory}. Most of the unsupervised instance-based approaches make the assumption of covariate-shift \cite{Bickel2009CovariateShift}.

\smallskip

\noindent \textbf{Importance weighting and deep learning}. This work is related to existing works in importance weighting using deep learning. Recently, Fang et al. \cite{Fang2020DeepImportanceWeighting} have developed a task-oriented sample-bias correction method where a KMM is performed at each batch in different depths of the neural network. In a different context from ours, Diesendruck et al. \cite{Diesendruck2019ImportanceWeightedGenerativeNetwork} have developed a sample bias correction method for deep generative models. In this approach, the MMD is also minimized by batch, however the weights are not parameterized but assumed to be known (e.g. a uniform proportion of classes is desired). Importance weighting methods have also been used along with deep feature transformation in partial domain adaptation \cite{Cao2018SAN4PDA}, i.e. when the number or the proportion of classes differ between targets and sources. In this category of methods, the output of a domain classifier network is often used to reweight the source instances. The domain classifier is either trained in parallel to the feature transformation \cite{Cao2019LearningToTransfer4PDA}, \cite{You2019UDA}, \cite{Zhang2018ImportanceWeightedAdversarial4PDA} or after it \cite{Park2020CalibrationCovariateShift}, \cite{wang2020TransferableCalibrationDA}. Other methods in this field consider the uncertainty of a task classifier to reweight both source and target instances during the feature transformation \cite{Guan2021UncertaintyAwareDA}, \cite{Wen2019BayesianUncertaintyWeightingDA}. These works are interesting from a computational point of view and may be seen as an alternative to MMD minimization. Finally, the weighting adversarial neural network (WANN) \cite{deMathelin2020WANN}, explicitly proposes to use a neural network to learn the source weights minimizing a distance between distribution called the $\mathcal{Y}$-discrepancy \cite{Mohri2010Ydiscrepancy}. Their approach, however, is developed in the supervised context and involves a task network fitted at the same time as the weights. Their approach is then deep learning specific. The present work generalizes this last approach as any estimator can be used once the weights are computed.

\section{Experiments}

We conduct the experiments on a synthetic dataset and $15$ UCI datasets\footnote{\url{https://archive.ics.uci.edu/ml/datasets.php}} \cite{Dua2019UCI} of various size and number of features. The experiments are conducted on a $3.3$Ghz computer with $64$G RAM and $24$ Cores. The source code of the experiments is available on GitHub\footnote{\url{https://github.com/antoinedemathelin/Importance-Weighting-Network}}.


\subsection{IWN Settings}

The purpose of IWN is to provide a simple and fast tool to perform importance weighting. We observe that the choice of network has little incidence on the learned weights (see Section \ref{impact}), we then arbitrarily choose a three layers neural network with $100$ neurons each and a ReLU activation. This architecture is used in all experiments without fine-tuning. We choose the Adam optimizer \cite{Kingma2014Adam}. The optimization parameters are also fixed for all experiments to a learning rate of $0.001$, a batch size of $256$ and a maximal number of iterations set to $5 \cdot 10^4$. Early stopping on the objective function is used, if the objective has not improved after $2 \cdot 10^4$ iterations, the learning is stopped.

We remind that the kernel bandwidth $\sigma$ used to compute the MMD is learned in the gradient descent (cf Algorithm \ref{algo-iwn}) and does not require a cross-validation process. The initial value of $\sigma$ is set to $0.1$ for all experiments.

\subsection{Competitors Settings}

We consider the following competitors which have already been introduced previously in this paper:

\begin{itemize}
    \item KMM \cite{Huang2007KMM}. We use a Gaussian kernel and the default optimization parameters $B=1000$ and $\epsilon = \sqrt{m-1} / \sqrt{m}$. The bandwidth $\sigma$ of the kernel is selected in the set $\{10^{(i-4)}\}_{i \in [|0, 8|]}$ with unsupervised cross-validation using the linear discrepancy \cite{Mansour2009DATheory}.
    \item KLIEP \cite{Sugiyama2007KLIEP}. We use a Gaussian kernel and a learning rate of $0.01$ with a maximum number of iterations of $1000$ as parameters for the gradient descent. These parameters have been selected to obtain an important decrease of the objective function with a fast convergence for most of the datasets. The bandwidth $\sigma$ of the kernel is selected in the set $\{10^{(i-4)}\}_{i \in [|0, 8|]}$ with the native Likelihood Cross-Validation (LCV) procedure of KLIEP.
    \item NNW \cite{loog2012NearestNeighborsWeighting}. The nearest neighbors are computed with the optimized NearestNeighbors algorithm of scikit-learn \cite{Pedregosa2011scikit-learn} which optimizes the computation approach between brute force and KD-Ball-Tree in function of the number of features and samples in the dataset. The Euclidean distance is used and the number of nearest neighbors for averaging is chosen in the set $\{ 1, 5, 10, 20, 50, 100 \}$ with unsupervised cross-validation using the linear discrepancy.
\end{itemize}

The implementation of the competitors are provided by the ADAPT library\footnote{\url{https://github.com/adapt-python/adapt}} \cite{Demathelin2021ADAPT}. The library also provides the metric for the cross-validation processes.

To offer the best chance to the competitors, the parameters selection with cross-validation are performed with parallel computing for KMM and NNW. For KLIEP, the parallel computing is not available in ADAPT and thus not used, this explains why its computational time is higher than the others.

\subsection{Synthetic dataset}

\begin{figure}[H]
	\begin{center}
		\includegraphics[width=0.99\textwidth]{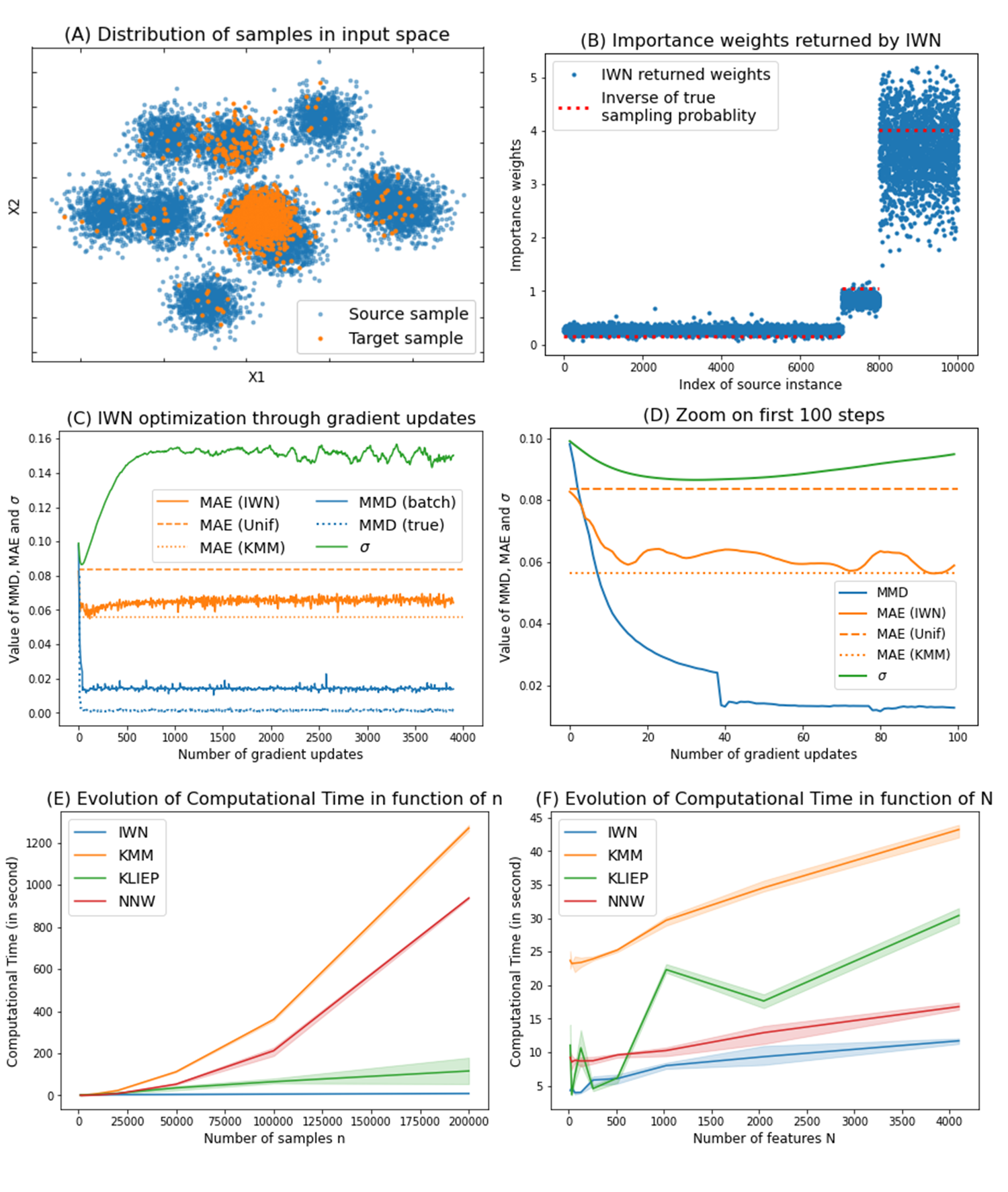}
		\caption{Visualization of the synthetic experiments. The corresponding experimental settings are: (A) $n=10000, N=2$, (B, C, D) $n=10000, N=32$, (E) $n \in [100, 2 \cdot 10^5], N=128$, (F) $n = 2 \cdot 10^4, N \in [16, 4096]$}
		\label{synth}
	\end{center}
\end{figure}

We consider the synthetic experiment, inspired from \cite{Cortes2014DAregression}, where $p_s(x)$ is a mixture of $M=10$ Gaussians, i.e. $p_s(x) = \sum_{k=1}^M \pi_k \, \mathcal{N}(\mu_k, 0.2)(x) $ where the centers $\mu_k \in \mathbb{R}^N$ are drawn according to the distribution $\mathcal{N}(0, 1)$ in $\mathbb{R}^N$, the ratios $\pi_k$ are set such that $\pi_k = 0.8/(M-1) \; \forall k \neq M$ and $\pi_M = 0.2$. The output variable is written $y = \beta_k^T x$ for any $x$ drawn according to the $k^{\text{th}}$ Gaussian. The coefficients $\beta_k \in \mathbb{R}^N$ are drawn according to the distribution $\mathcal{N}(0, 1)$ in $\mathbb{R}^N$. The target distribution $p_t(x)$ is drawn according to the same mixture of Gaussians but with ratios $\pi'_k = 0.1 / (M-2) \; \forall k < M-1, \pi'_{M-1} = 0.1$ and $\pi'_M = 0.8$. We suppose that the learner has access to a sample of size $n$ of source labeled instance $\{(x_i, y_i)\}_{1 \leq i \leq n}$ drawn according to the source distribution $p_s(x, y)$ on $\mathbb{R}^N$ and an unlabeled set of size $n$, $\{x'_j \}_{1 \leq j \leq n}$ drawn according to the target distribution $p_t(x)$ on $\mathbb{R}^N$. An illustration of the problem for dimension $N=2$ is given in Figure \ref{synth}.A.

We conduct several experiments on this synthetic dataset. First, we fit IWN for the setting $N=32, n=10000$. We make $4000$ batch updates with batch size $256$, at each of them, we use the weights $w_i$ returned by the weighting network to fit a weighted Ridge regression model of parameters $\beta$ on the set $\{(x_i, y_i, w_i)\}_{1 \leq i \leq n}$, we then record the mean absolute error (MAE) of this model on the target dataset: $\frac{1}{n} \sum_i |\beta^T x'_i - y'_i|$. We also record, at each batch, the computed MMD on the batch and the "true" MMD computed with the whole samples. We also record the current value of the parameter $\sigma$ which is also updated during the optimization (see Section \ref{iwn}). We report the results of this experiment on Figure \ref{synth}.B, \ref{synth}.C, \ref{synth}.D. We first represent the final importance weights returned by the weighting network at the end of the $4000$ iterations in Figure \ref{synth}.B. As we can observe, the learned weights are very close to the true sampling probability. On Figure \ref{synth}.C, we report the evolution of the recorded MAE in plain orange, the batch and true MMD in blue and the value of $\sigma$ in green. We also report, for comparison with "MAE (IWN)", the MAE of a Ridge model fitted with uniform weights: "MAE (Unif)" and the MAE of the model fitted with the weights obtained using KMM: "MAE (KMM)". We observe that the importance weighting produced by IWN helps to learn the task on the target domain as the error decreases of $20\%$ compared to the error of the model fitted with uniform weights. Concerning the recorded MMD, we observe that both the batch and "true" MMD decrease very fast, but an offset remains between the two due to the estimation error made with finite samples. We then see on the zoom of Figure \ref{synth}.D that the error decreases very fast as well as the MMD. After $100$ iterations, the MMD is minimized and the target error of IWN is on the same level than the error produced by KMM. We notice that the MAE increases a little after some iterations which may indicates some overfitting effect of the weighting network. This observation argues for the use of early stopping based on the evolution of the MMD.

\noindent We finally remark, in Figure \ref{synth}.C, that the $\sigma$ parameter becomes relatively stable around $0.15$ after some iterations which comforts the idea that the parameter can be efficiently set during the optimization thanks to the adversarial learning (see Section \ref{iwn}). We could however argue, that the increase of the target error after the $100^{\text{th}}$ iteration is correlated to the increase of $\sigma$ from $0.1$ to $0.15$. This is a plausible explanation as $\sigma$ is updated in order to make the MMD more discriminative which may provoke overfitting. This is a limitation of the proposed approach which we propose to study in future work.

We then conduct several experiments to observe the evolution of the computational time in function of the sample size and the number of features. We first fix the number of features to $N = 128$ and vary the number of samples from $100$ to $2 \cdot 10^5$ and report the computational time in Figure \ref{synth}.E. Then, we fix the number of samples to $n=20000$ and vary $N$ from $16$ to $4096$ and report the results in Figure \ref{synth}.F. We observe, on these two Figures, that for the number of samples has a stronger impact on the computational time than the number of features. The two methods KMM and NNW have a quadratic complexity $\mathcal{O}(n^2)$ which is well reflected in Figure \ref{synth}.E. The computational time of KLIEP evolves linearly due to the fixed number of target centers considered. Figure \ref{synth}.E clearly demonstrates the computational supremacy of IWN compared to the other methods. It should be pointed out that the quality of the importance weighting is similar between methods, the corresponding scores are reported in appendix.

\subsection{UCI datasets}

\begin{figure}[H]
	\begin{center}
		\includegraphics[width=0.90\textwidth]{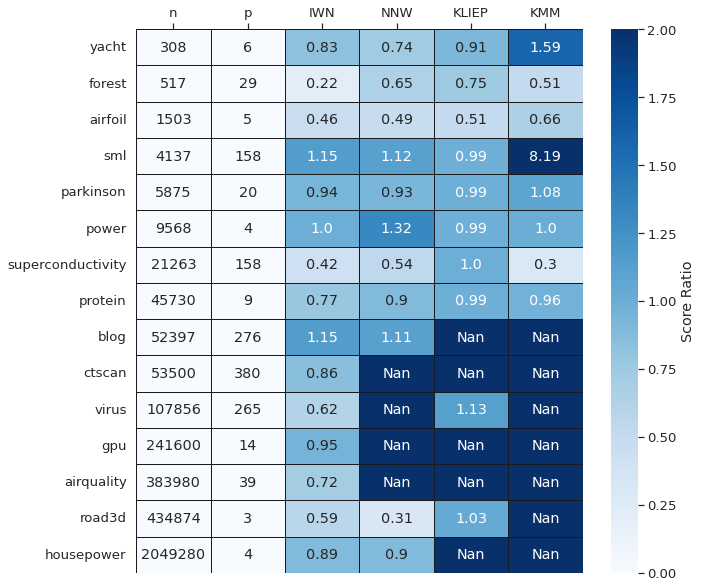}
		\caption{Score ratio for the experiments with the sample bias applied on input features. In each column, the values correspond to the ratios between the mean absolute error (MAE) of the corresponding methods (IWN, NNW, KLIEP and KMM) and the MAE of the Uniform Weighting approach where no reweighting is performed. The MAEs are computed on the target dataset with a Ridge model fitted on the reweighted source data. The two first columns $n$ and $p$ are respectively the number of sample and number of features of the dataset. The lightest colors correspond to the best ratios. The experiment stopped after 500 seconds are marked with Nan.}
		\label{bias-X-score}
	\end{center}
\end{figure}

\begin{figure}[H]
	\begin{center}
		\includegraphics[width=0.95\textwidth]{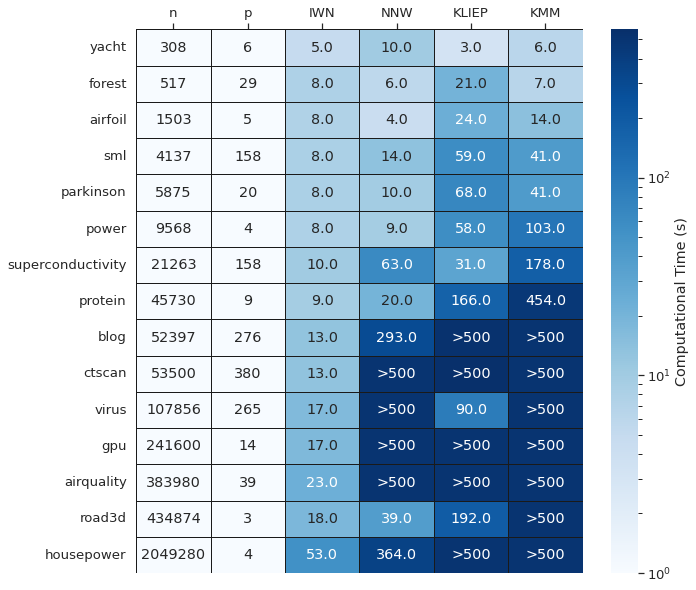}
		\caption{Computational times (in second) for the experiments with the sample bias applied on input features. The two first columns $n$ and $p$ are respectively the number of sample and number of features of the dataset. The computational time are given in second. The lightest colors correspond to the lowest computational time. The experiment stopped after 500 seconds are marked (>500).}
		\label{bias-X-times}
	\end{center}
\end{figure}

We perform the experiments on several UCI datasets \cite{Dua2019UCI} with different sizes and dimensions. We record the computational time used by each importance weighting method for computing the source importance weights (cf Figure \ref{bias-X-times}). For each dataset, the score is computed with a Ridge model fitted with the importance weights and without importance weighting. The ratio between the two scores is reported in Figure \ref{bias-X-score}.

To evaluate the importance weighting methods we consider different kind of sample bias following the setting of \cite{Huang2007KMM}:

\begin{itemize}
    \item Sample bias on the input features: the source training set is biased on the input features $X$ with a gaussian weighting on the first component of the PCA of mean $m + (\mu - m)/3$ and standard deviation $(\mu - m)/8$ with $m, \mu$ the respective minimum and mean of the first PCA component.
    \item Sample bias on the output features: the source training set is biased on the output features $y$ with a weighting on the first component of the PCA defined as $\exp(3(y_{1}-1))/(1 + \exp(3(y_{1}-1)))$ with $y_1$ the first PCA component of $y$.
\end{itemize}

The training set is built by taking $n$ data with replacement using the sampling bias as probabilities of selection. The target set is the original dataset without selection bias. For each experiment, we apply the following preprocessing: standard scaling of the numerical inputs (using the mean and standard deviation of the unbiased inputs) and one-hot-encoding of the categorical inputs. The dimension $p$ of the input space corresponds to the dimension after preprocessing.

To evaluate the weighting scheme of each method, we fit a Ridge model with trade-off parameters $\alpha$ selected by leave-one-out process between values $\{10^{(i-4)}\}_{i \in [|0, 8|]}$. For regression datasets, we compute the mean absolute error (MAE) on the target testing set (the original dataset without bias). We then fit another Ridge model with the uniformly weighted biased source data and compute the MAE on the target set. We can then compute a score ratio for each method as follows:

\begin{equation}
\label{ratio}
    \text{Score Ratio} = \frac{ \sum_i | \beta_{IW}^T x'_i - y'_i |}{ \sum_i | \beta_{Unif}^T x'_i - y'_i |}
\end{equation}

where $\beta_{IW}, \beta_{Unif}$ are respectively the coefficients of the Ridge models fitted with the importance weights and the uniform weights.

We repeat each experiment $10$ times and report the results of the experiments with the sample bias on the input features in Figures \ref{bias-X-score} and \ref{bias-X-times}, the standard deviation over the $10$ repetitions are given in appendix. The first Figure presents the score ratios computed with Eq (\ref{ratio}). We observe that the quality of the weighting scheme provided by IWN is competitive with other methods. IWN is in the top 2 best ratios for all experiments except one. Even more impressive are the computational time of IWN compared to other methods (Figure \ref{bias-X-times}), IWN provides a fast computation of no more than 25 seconds for samples of size $< 5 \times 10^5$. Whereas the other methods failed to provide importance weights in a reasonable time for samples above $5 \times 10^4$ samples and $>100$ features. The same observations are made on Table \ref{bias_y} which reports the summary results of the experiments conducted on the same datasets with a sample bias on the outputs. 

\begin{table}[h]
\centering
\begin{tabular}[width=\textwidth]{c|c|c|c}
\toprule
      method & Avg Score Ratio & Avg Rank & Avg Comp. Time (in sec.) \\
      \midrule
IWN   & 0.9            & \textbf{2.0}      & \textbf{8.06}                    \\
NNW   & 0.92            & 2.37     & 182.81                   \\
KLIEP & 1.03            & 2.93     & 223.16                   \\
KMM   & \textbf{0.87}            & 2.7      & 290.01 \\
\bottomrule
\end{tabular}

\caption{Summary of the results of the output sample bias experiments (extensive results are reported in appendix).}
\label{bias_y}
\end{table}

\subsection{Impact of Network Architecture and Batch Size}
\label{impact}

Finally, we study the impact of the network architecture and the batch size on the solution of IWN. We conduct the experiments on the CTscan dataset\footnote{\url{https://archive.ics.uci.edu/ml/datasets/Relative+location+of+CT+slices+on+axial+axis}} \cite{Graf2011CTscan} biased through the sample bias on the input features described previously. First, we fix the batch size to $256$ and vary the number of hidden layers of the weighting network from $0$ to $4$ and the number of neurons per layer between $[10, 100, 300]$. Then, we fix the number of hidden layers to $3$ and the number of neurons to $100$ and vary the batch size on a geometric scale of ratio $4$ from $16$ to $4096$. We repeat each experiment $10$ times and report the means and standard deviations of the scores in Table \ref{ablation}. We observe that the architecture of the network has little impact on the score, however, we observe a slight improvement of the score between the simplest architectures and the more complex ones ($0.88$ for $0$ hidden layer and $0.85$ for $4$ layers and $300$ neurons). The impact of the batch size is more significant on the performance of IWN, enlarging the batches produce better corrections of sample bias. This is due to the more accurate estimation of the MMD made with larger batch. However, increasing the batch size comes with an increase of the computational time as shown in the last column of Table \ref{ablation}.

\begin{table}[h]
\centering
\begin{tabular}[width=\textwidth]{c|>{\centering\arraybackslash}p{1.6cm}|>{\centering\arraybackslash}p{1.6cm}|>{\centering\arraybackslash}p{1.6cm}||>{\centering\arraybackslash}p{0.8cm}|>{\centering\arraybackslash}p{1.6cm}|c}
\toprule
\multicolumn{4}{c||}{\textbf{Neural Network Architecture}} & \multicolumn{3}{c}{\textbf{Batch Size}} \\
\hline
\backslashbox{Layers}{Neurons}  & 10          & 100         & 300      & Size & Score       & Time (s)       \\
\hline
0 & 0.88 (0.02) & 0.88 (0.02) & 0.88 (0.02)& 16   & 0.96 (0.01) & 9.7 (8.2)   \\
1 & 0.86 (0.02) & 0.86 (0.02) & 0.85 (0.02)& 64   & 0.90 (0.01)  & 6.6 (0.2)   \\
2 & 0.86 (0.02) & 0.86 (0.02) & 0.85 (0.02)& 256  & 0.85 (0.02) & 7.4 (0.3)   \\
3 & 0.86 (0.02) & 0.85 (0.02) & 0.85 (0.02)& 1024 & 0.82 (0.03) & 12.1 (0.1)  \\
4 & 0.86 (0.02) & 0.85 (0.02) & 0.85 (0.02) &4096 & 0.83 (0.04) & 117.9 (1.3) \\
\bottomrule
\end{tabular}
\caption{Summary of the study on the impact of weighting network architecture and batch size. Standard deviation over the $10$ repetitions are given in brackets.}
\label{ablation}
\end{table}

\section{Conclusion}

This work introduces a novel algorithm for importance weighting called Importance Weighting Network and shows that sample biases can be efficiently corrected by fitting a weighting neural network with the MMD as loss function. This approach appears to provide very competitive results with state-of-the-art instance-based domain adaptation methods for a minimal cost in term of computational time.


\end{document}